\documentclass[
compsoc,journal]{IEEEtran}

\usepackage{algorithm}
\usepackage{algorithmic}
\usepackage{float}
\usepackage{graphicx}
\usepackage{hyperref}

\begin{document}
%

\title{Image Segmentation with Superpixel-based Covariance Descriptors in Low-Rank Representation}

\author{\IEEEauthorblockN{Xianbin Gu, 
Jeremiah D. Deng,~\IEEEmembership{Member,~IEEE}, and
Martin K. Purvis}
\IEEEauthorblockA{
Department of Information Science, University of Otago, Dunedin 9054, New Zealand}
\thanks{
Corresponding author: Jeremiah D. Deng (email: jeremiah.deng@otago.ac.nz).}}

\maketitle

\begin{abstract}
This paper investigates the problem of image segmentation using superpixels. We propose two approaches to enhance the discriminative ability of the superpixel's covariance descriptors. In the first one, we employ the Log-Euclidean distance as the metric on the covariance manifolds, and then use the RBF kernel to measure the similarities between covariance descriptors. The second method is focused on extracting the subspace structure of the set of covariance descriptors by extending a low rank representation algorithm on to the covariance manifolds. Experiments are carried out with the Berkly Segmentation Dataset, and compared with the state-of-the-art segmentation algorithms, both methods are competitive.

\end{abstract}




\section{Introduction}
Covariance matrix is widely used in image segmentation with whom those pixel-wise features, like color, gradient, etc., are assembled together into one symmetric positive definite matrix ($Sym^+_d$) named covariance descriptor. Research shows that such descriptor is a feature that highly discriminative for many image processing tasks, such as image classification, segmentation, and object recognition~\cite{gu2014improving,harandi2012kernel,tuzel2006region,yao2007fast}.  

In image segmentation, the covariance descriptor is often built on some basic features of a group of pixels, like color, intensity, positions, or gradients, etc. And different combinations of basic features bring different covariance descriptors, which often influence the algorithms' performance. However, the research about how to construct covariance descriptors for a specific task is insufficient, and the relations between basic feature combinations and performance of the algorithms are still unclear. 

Generally, researchers construct covariance descriptors by trying different basic feature combinations and taking the one that gives the best performance~\cite{habibouglu2012covariance,kviatkovsky2013color}, i.e. by an empirical way. This may be handy for practice but lack of theory support. 

Fortunately, there are some other options, which can enhance the discriminative ability of covariance descriptors without repeatedly testing the combinations of the features. These  methods are based on two facts. First, in the view of differential geometry, the covariance descriptors are symmetric positive definite matrices lying on a convex cone in the Euclidean space, i.e. they are the points on some $Sym_d^+$ manifolds. So, a proper defined metric on the $Sym_d^+$ will improve the performance of covariance descriptors. Secondly, there are always some correlations between the the basic features that built up the covariance descriptor, which may bring noises. Thus, removing the noise in covariance descriptors is also beneficial. 

To this end, we study the RBF kernel and Low Rank Representation algorithm on the $Sym_d^+$ manifolds and compare them with the state-of-the-art methods by experimenting on different covariance descriptors extracted from superpixels.

The rest of the paper is organized as follows. $Section$~\ref{preliminary} presents a brief review of the algorithms and concepts that are necessary for this paper. $Section$~\ref{similarity} discusses different methods to measure the similarity of covariance descriptors. $Section$~\ref{experiment} is the comparison between the algorithms, and $Section$~\ref{conclusion} gives the conclusion.  

\section{Background}
\label{preliminary}
\subsection{Manifold, covariance descriptor and multi-collinearity}
A manifold $\mathcal{M}$ of dimension $d$ is a topological space that is locally homeomorphic to open subset of the Euclidean space $R^d$~\cite{xie2013nonlinear}. For analysis on $\mathcal{M}$, there usually have two options. One is by embedding $\mathcal{M}$ into $R^d$ so that to create a linear structure and the methods from Euclidean space can be directly applied. The alternative is concentrated on the intrinsic properties of $\mathcal{M}$, i.e. define a properly intrinsic metric of $\mathcal{M}$ and  launch the analysis on the true structure of $\mathcal{M}$. The first method is more intuitive but less accurate because the extrinsic metric may not align with the intrinsic properties of the manifold. While the second method is able to represent the manifold structure precisely, but for analysis,  it is not as friendly as in the Euclidean space. 

Let $\textbf{F} = (\textbf{f}_1,...,\textbf{f}_n)^T$ be a feature array, where $\textbf{f}_i$ is a vector whose entries are the observations of the $i$-th feature. A covariance descriptor is the covariance matrix of $\textbf{F}$, which is defined as,
\begin{equation}
\label{cov-matrix}
cov(\textbf{F})=\left[E((\textbf{f}_i-\mu_i)^T(\textbf{f}_j-\mu_j))\right]_{n\times n}, 
\end{equation}
where $\mu_i$ is the mean of the $i$-$th$ feature $\textbf{f}_i$, and
$[\cdot]_{n\times n}$ indicates an $n\times n$ matrix whose entry at position $(i,j)$ is represented by the ``$\cdot$'' . Apparently, different sets of $\textbf{f}_i$ generate different $cov(\textbf{F})$, which makes the performance vary.

Collinearity (or multi-collinearity) is a term from statistics, which refers a linear association between two (or more) variables. Specifically, given a feature array $\textbf{F}$, if there exists a set of not-all-zero scalar $\lambda_1,...,\lambda_n$ that makes the following equation holds,
\begin{equation}
\label{multi-collinearity}
\lambda_1 \textbf{f}_1+\lambda_2\textbf{f}_2+\cdots+\lambda_n\textbf{f}_n + u =0
\end{equation}
 If $u=0$, $\textbf{F}$ is perfect multi-collinearity; while if $u\sim N(0,\sigma)$, $\textbf{F}$ is nearly multi-collinearity. This multi-collinearity phenomenon is common in image segmentation because the variables in the feature array $\textbf{F}$, like gradient, are computed from other containing variables (i.e. inclusion). If we consider $Eq.$(\ref{multi-collinearity}) and $Eq.$(\ref{cov-matrix}) simultaneously, it is easy to see the inclusion may produce redundant entries and noises in $cov(\textbf{F})$.

\subsection{Segmentation with superpixels}
A cluster of pixels is called superpixel for whom its members are more similar between each other than those nonmembers. In ~\cite{cheng2011multi,li2012segmentation}, the authors have shown that image segmentation benefits from using superpixels. There are three advantages for superpixel based segmentation. Firstly, they dramatically reduce computation cost by representing pixels inside one superpixel as a whole. Secondly, regional features can be extracted from superpixels, which are more discriminative than pixel-wise features in many vision tasks. Thirdly, multi-scale techniques can be applied when considering different parameter settings (or, different algorithms) for superpixel generating as different scales. 

In this paper, the segmentation is performed by a spectral clustering algorithm proposed in~\cite{li2012segmentation}, which takes advantage of the superpixels. Briefly, this algorithm mainly contains three parts. The first one is superpixel generation. Similar to~\cite{li2012segmentation,gu2014improving,wang2013sparse}, we produce superpixels by Mean Shift algorithm~\cite{comaniciu2002mean} and Felzenszwalb-Huttenlocher algorithm~\cite{felzenszwalb2004efficient} with a few sets of parameters. The second is  graph construction. We construct a bipartite graph on the pixels and superpixels, and the superpixels are represented by covariance descriptors. We adopt different ways to measure the similarities of the superpixels, and the results are reported in $Section$~\ref{conclusion}. The third part is spectral clustering. We use $T$-$cut$~\cite{li2012segmentation} to reduce the computational cost. The framework is shown in Alg.\ref{alg1}. 
\begin{algorithm}
\caption{Superpixel-based Segmentation}
\label{alg1}
\begin{algorithmic}[1]
\REQUIRE An image $M$ and the number of segmentation $k$.
\ENSURE A $k$-way segmentation of $M$.
\STATE Create over segmentation of $M$ by superpixel algorithms;
\STATE Construct bipartite graph $G(X,Y,B)$ (refer to $Section$~\ref{construction_graph});
\STATE Compute weights on the edges by some similarity measure (refer to Alg.\ref{alg:RBFLE} and Alg.\ref{alg:LRR}); 
\STATE Partition $G$ by $T$-$cut$.  
\end{algorithmic}
\end{algorithm}

\section{Similarity measurement}
\label{similarity}
\subsection{Construction of bipartite graph}
\label{construction_graph}
Let $P=\{p_1,...,p_n\}$ denotes the set of pixels of a given images and $\mathcal{S}$ be a collection $\{S_i\}$, where $S_i = \{s^{(i)}_t\}$ is a set of superpixels. A given bipartite graph $G(X,Y,B)$ is built in this way: set $X=P\cup \mathcal{S}$, $Y=\mathcal{S}$ and let $B=E_1\cup E_2$ be the edges, where $E_1$ and $E_2$ represent the edges in every vertex pair $(p_k, s_t^{(i)})$ and $(s^{(i)}_k, s^{(i)}_l)$ respectively. A number of methods have been proposed to  the similarities between superpixels (i.e.,the weights on the edges in $E_2$). Some are based on geometry constrains~\cite{li2012segmentation}, some are based on encoding techniques~\cite{wang2013sparse}. While in~\cite{gu2014improving}, a covariance descriptor based method was proposed, which employed color and a covariance matrix of the color value $[R,G,B]$ to assess the similarities in superpixels. 

However, it is still necessary to explore the mechanism of making use of covariance descriptors. Our first thinking is to use the geodesics distance measuring the distance between the covariance descriptors. Secondly, we note that the multi-collinearity is almost inevitable when building the covariance descriptors, this encourages us to use the low rank representation algorithm.

\subsection{The RBF kernel}

A dataset of $d\times d$ covariance matrices is lying on a manifold embedding in $d^2$-dimensional Euclidean space because the space of $d\times d$ covariance matrices is a convex cone in the $d^2$-dimensional Euclidean space. 

It has been proofed that the geometry of the space $Sym_d^+$ can be well explained with a Riemannian metric which induced an infinite distance between an $Sym_d^+$ matrix and a non-$Sym_d^+$ matrix~\cite{arsigny2005fast,pennec2006riemannian,jayasumana2015kernel}. But different to~\cite{gu2014improving}, we use the Log-Euclidean distance, a geodesic distance, which makes it possible to embed the manifold into $\mathcal{RKHS}$ with RBF kernel~\cite{jayasumana2015kernel}.

The Log-Euclidean distance between covariance matrix $X_i$ and $X_j$ is defined as $d_{LE}(X_i,X_j):=\|Log(X_i)-Log(X_j)\|_F$. Respectively, the RBF kernel can be rewritten as,
\begin{equation}
k_{LE}:(X\times X)\rightarrow R: k_{LE}(X_i,X_j):=exp(-\sigma d^2_{LE}(X_i,X_j))
\end{equation}
where $\sigma>0$ is a scale parameter. The similarity matrix is defined as $[B]_{ij}:=(k_{LE}(X_i,X_j))$. Details are as Alg.\ref {alg:RBFLE}.
\begin{algorithm}[!htbp]
\caption{Construct the Graph via RBF kernel}
\label{alg:RBFLE}
\begin{algorithmic}[1]
\REQUIRE The Graph $G(X,Y,B)$; parameter $\alpha$ and $\sigma$ .
\ENSURE A weighted Graph $G(X,Y,\tilde{B})$.
\FORALL{$e$ in $B$} 
\IF {$e\in E_2$ }
\STATE{$e=exp(-\sigma d^2_{LE}(s_k^{(i)}, s_l^{(i)}))$}
\ELSE
\STATE{$e = \alpha$}
\ENDIF 
\ENDFOR
\end{algorithmic}
\end{algorithm}

Theoretically, the kernel method is more efficient, but it may not hold in our case. We give some possible explanation in $Section$~\ref{conclusion}.  

\subsection{Low rank representation for $Sym_d^+$}
The low rank representation (LRR) algorithm~\cite{candes2011robust,wright2009robust} is proposed to remove the redundancy and noises in the dataset. For a given dataset $X$, the LRR algorithm finds a matrix $Z$, called low rank representation of $X$, such that $X=XZ + E$ holds, where $E$ represents the noises. The performance of LRR is promising when $X$ is a set of points in Euclidean space~\cite{ganesh2009fast,chen2014robust,liu2011latent}, and recently, it has been extended to data sets lying on manifolds~\cite{fu2015low,wang2015kernelized,wang2015low}.

Let $\mathcal{X}$ be a 3-order tensor, which is stacked from covariance matrices $(X_i)_{d\times d}$, $i=1,2,...,n$. By embedding $\mathcal{X}$ into the $d^2$-dimensional Euclidean space, our LRR model is set as follows,
\begin{equation}
\label{LRR}
\begin{array}{ll}
&\min_{E,Z} {\|E\|_F^2 + \lambda\|Z\|_{\ast}},\\
&s.t. \quad \mathcal{X} = \mathcal{X}_{\times _3}Z+E,
\end{array}
\end{equation}
where $\|\cdot\|_F$ is the {\em Frobenius} norm; $\|\cdot\|_{\ast}$ is the {\em nuclear} norm; $\lambda$ is the balance parameter; $\times_3$ means mode-3 multiplication of a tensor and matrix~\cite{kolda2009tensor}. Eq.(\ref{LRR}) can be solved via Augment Lagrangian Multiplier (ALM)~\cite{lin2010augmented}. Details are in Appendix. 

Let $\Delta$ be a symmetric matrix, whose entries are $\Delta_{ij}=\Delta_{ji}=tr(X_iX_j)$, we can obtain a solution of Eq.(\ref{LRR}) by iteratively updating the following variables,
\begin{equation}
J = \Theta(Z+\frac{Y}{\mu}), 
\end{equation}
and,
\begin{equation}
Z = (\lambda\mu J -\lambda Y +2\Delta)(2\Delta +\lambda\mu I)^{-1},
\end{equation}
where $\lambda$ and $\mu$ are pre-setting parameters, $Y$ is the Lagrange coefficient, $\Theta(\cdot)$ is the singular value thresholding operator~\cite{cai2010singular}.

Since the coefficient matrix $Z$ contains the subspace information of the dataset, it is reasonable to define the similarity matrix as $[B]_{ij}:=([\widetilde{U}\widetilde{U}^T]_{ij})^2$ for spectral clustering, where $\widetilde{U}$ is the row-normalized singular vector of $Z$~\cite{liu2013robust}.

\section{Experiments}
\label{experiment}
The experiments are set to compare the effects of different similarity measure algorithms of covariance descriptors. We construct the covariance descriptors with three different feature vectors named as CovI, CovII, CovIII respectively,
\begin{itemize}
\item 
CovI: $[R,G,B]$,
\item 
CovII:$[R,G,B, I,\frac{\partial I}{\partial x}, \frac{\partial I}{\partial y},\frac{\partial I}{\partial^2 x},\frac{\partial I}{\partial^2 y}]$, 
\item  
CovIII: $[R,G,B,\frac{\partial R}{\partial x},\frac{\partial R}{\partial y},\frac{\partial G}{\partial x},\frac{\partial G}{\partial y},\frac{\partial B}{\partial x}$,\\
$\frac{\partial B}{\partial y},\frac{\partial R}{\partial^2 x},\frac{\partial R}{\partial^2 y},\frac{\partial G}{\partial^2 x},\frac{\partial G}{\partial^2 y},\frac{\partial B}{\partial^2 x},\frac{\partial B}{\partial^2 y}]$.
\end{itemize}

From CovI to CovIII, the dimensionality of the covariance descriptor is increasing. For example, CovI contains the patterns in the $R$, $G$, $B$ channels, while in CovIII, the patterns of their derivatives are also included. This means the covariance descriptors become more discriminative. But, since the partial derivatives are directly computed from other contained features, the tendencies of multi-collinearity are also growing.

The methods for similarity measurement include RBF kernel with Log-Euclidean distance (RBFLE) and Low Rank Representation (LRR).

All experiments are done with the Berkly Segmentation Dataset, a standard benchmark image segmentation dataset, which includes 300 natural images selected from diverse scene categories~\cite{arbelaez2011contour}. Besides, it also provide a number of human-annotated ground-truth descriptions for each image. In our experiments, every image is partitioned into $K$ regions with $K\in [2,40]$. And, the reported evaluation results are based on the $K$ that provides the best performance of the algorithms. 

\begin{algorithm}[!htbp]
\caption{Construct the Graph via LRR}
\label{alg:LRR}
\begin{algorithmic}[1]
\REQUIRE A collection of covariance matrix  $\mathcal{S}$; parameter $\lambda$.
\ENSURE A weighted similarity matrix $\tilde{B}$.
\FOR {$S_i=\{s_t^{(i)}\}_{t=1}^{n}$ in $\mathcal{S}$} 
\STATE Compute $Z_i$ of $S_i$:
\STATE Initialize: $J=Z_i=0$, $Y=0$, $\mu=10^{-6}$, $\mu_{max}=10^{10}$, $\rho=1.9$, $\epsilon = 10^{-8}$;
\FOR{i=1:n}
\FOR{j=1:n}
\STATE $\Delta_{ij}=tr[(s_t^{(i)}s_t^{(j)})]$
\ENDFOR
\ENDFOR
\WHILE{not converged}
\STATE Fix $Z_i$ and update $J$ by\\
$J \leftarrow \Theta(Z_i+\frac{Y}{\mu})$;
\STATE Fix $J$ and update $Z$ by\\
$Z \leftarrow (\lambda\mu J -\lambda Y +2\Delta)(2\Delta +\lambda\mu I)^{-1}$
\STATE Check convergence,
\IF{$\|Z_i-J\|_F<\epsilon$}
\STATE break
\ELSE
\STATE Update $Y$ and $\mu$,\\
$Y\leftarrow Y + \mu(Z_i-J)$\\
$\mu\leftarrow min(\rho\mu, \mu_{max})$
\ENDIF
\ENDWHILE
\STATE Compute Singular Value Decomposition of $Z_i$,\\
$Z_i = U_i\Sigma_i V_i^T$
\STATE Compute the weights on $e\in\{(s_k^{(i)},s_l^{(i)})\}_{k,l=1}^{n}\subset{B}$ by\\ 
$(\tilde{U_i}*\tilde{U_i}^t)^2$, where $\tilde{U_i}$ is the $U_i$ normalized in rows.

\ENDFOR
\end{algorithmic}
\end{algorithm}

\subsection{The settings}
The same as in~\cite{gu2014improving,wang2013sparse,li2012segmentation}, the superpixels are generated by the Mean Shift algorithm with 3 different sets of parameters and the Felzenszwalb-Huttenlocher algorithm with 2 different sets of parameters; the weights on $E_1$ are fixed to $1\times 10^{-3}$. For the RBF-LE, parameter $\sigma$ is set as 20. In the LRR model, parameter $\lambda$ is sensitive to noise level~\cite{liu2013robust}. Since the BSDS contains images from different categories, which indicates the noise levels are different between images, we tuned $\lambda$ for every image by a grid research in  $\{1, 0.1, 0.01, 0.001\}$; a higher noise level is associated with a greater $\lambda$. In addition, we apply ``$k$-nearest neighbor'' to refine the similarity graph with $k=1$.

\subsection{The evaluation}
The performance of the algorithms is assessed by four popular segmentation evaluation methods, which includes the Probabilistic Rand Index (PRI)~\cite{unnikrishnan2007toward}, the Variation of Information (VoI)~\cite{meila:2005comparing}, the Global Consistency Error (GCE)~\cite{martin2001database}, and the Boundary Displacement Error (BDE)~\cite{freixenet2002yet}. 

PRI is a nonparametric test, which measures the probability of an arbitrary pair of samples being labeled consistently in the two segmentations. A higher PRI value means better segmentation.

The VoI is a metric that relates to the conditional entropies between the class label distribution. It measures the sum of information loss and information gain between the two partitions, so it roughly measures the extent to which one clustering can explain the other. A lower VoI value indicates better segmentation result. 

The GCE measures the difference between two regions that contain the same pixel in different segmentations. Particularly, this metric compensates for the difference in granularity. For GCE values, being close to $0$ implies a good segmentation.

The BDE measures the average displacement error of boundary pixels between two segmentations. The error of one boundary pixel is defined as the distance between the pixel and the closest pixel in the other boundary image. A lower BDE value means less deviation between the segmentation and ground truth.

In addition, we rank the algorithms on each index and give the average rank (Avg.R, lower is better) of each algorithm, which gives a straightforward comparison between the methods.

\subsection{Results} 
The evaluations of the algorithms are listed in the following tables. The results the state-of-art algorithms in~\cite{gu2014improving,wang2013sparse,li2012segmentation} (i.e. SAS, $\ell_0$-sparse, col+CovI) are also listed for reference. 

Table.\ref{tab:res-k-varying} shows the best results of the algorithms proposed in this paper together with those reported in other papers. For CovII+LRR, the PRI is the highest and the rest index values are very close to the other algorithms.
\begin{table}[htbp]
\caption{Performance of different algorithms}
\label{tab:res-k-varying}
\begin{tabular}{p{2.15cm}|p{0.78cm}p{0.78cm}p{0.78cm}p{0.78cm}p{0.78cm}}
\hline
{Algorithms}
&PRI&VoI&GCE&BDE&Avg.R\\
\hline\hline
SAS~\cite{li2012segmentation}&0.8319&1.6849&\textbf{0.1779}&11.2900&\ 2.5\\
\hline
$\ell_0$-sparse~\cite{wang2013sparse}&0.8355&1.9935&0.2297&\textbf{11.1955}&\ 3.75\\
\hline
Col+CovI~\cite{gu2014improving}&0.8495&\textbf{1.6260}&0.1785&12.3034&\ 2.25\\
\hline
CovII+LRR&\textbf{0.8499}&1.7418 &0.1915 &12.7635&\ 3 \\
\hline
CovIII+RBFLE &0.8397 &1.9026 &0.2103 &11.5557&\ 3.5\\
\hline
\end{tabular}
\end{table}
Table.\ref{tab:RBFLE-LRR} demonstrates the results from RBFLE and LRR. The performance of LRR is overwhelming, which indicates noises reduction inside the covariance descriptor benefits the segmentation results. For RBFLE, the performance is inferior even the similarities are measured by geodesic metric and kernel method. Because the inner noises (due to multi-collinearity) of the covariance descriptors may distort the true data values.
\begin{table}[H]
\caption{Performance of RBFLE and LRR with different covariance descriptors}
\label{tab:RBFLE-LRR}
\begin{tabular}{p{2.6cm}|p{0.95cm}p{0.95cm}p{0.95cm}p{0.95cm}}
\hline
{Algorithms}
&PRI&VoI&GCE&BDE\\
\hline\hline
CovI+RBFLE&0.8349 &2.0148&0.2218 &\textbf{12.5276} \\
\hline
CovI+LRR&\textbf{0.8454}& \textbf{1.7564}&\textbf{0.1885}&13.0427\\
\hline\hline
CovII+RBFLE&0.8372 &1.9466 &0.2148 &\textbf{11.6658}\\
\hline
CovII+LRR&\textbf{0.8499}&\textbf{1.7418} &\textbf{0.1915}&12.7635 \\
\hline\hline
CovIII+RBFLE &0.8397 &1.9026 &0.2103 &\textbf{11.5557}\\
\hline
CovIII+LRR &\textbf{0.8451}&\textbf{1.7698} &\textbf{0.1932}&12.4837\\
\hline
\end{tabular}
\end{table}

Moreover from Table.\ref{tab:RBFLE-LRR}, we notice that the performance of RBFLE goes up with the growing of the entries in the covariance descriptor while the result of LRR varies. One possible reason is that we use extrinsic metric in the LRR algorithm, which may influence the performance. 
\begin{figure}[H]
\begin{minipage}[b]{0.49\linewidth}
  \centerline{\includegraphics[width=\textwidth]{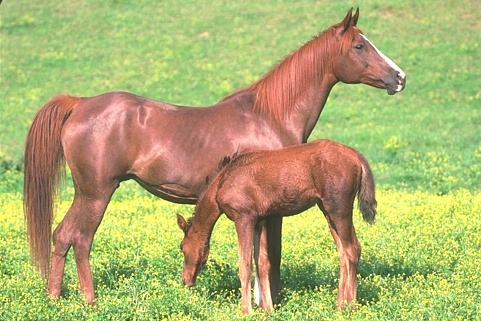}}
\end{minipage}
\begin{minipage}[b]{0.49\linewidth}
  \centerline{\includegraphics[width=\textwidth]{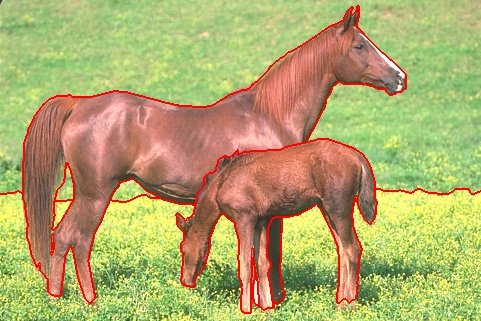}}
\end{minipage}
\begin{minipage}[b]{0.49\linewidth}
  \centerline{(a)}
\end{minipage}
\begin{minipage}[b]{0.49\linewidth}
  \centerline{(b)}
\end{minipage}
\begin{minipage}[b]{0.49\linewidth}
  \centerline{\includegraphics[width=\textwidth]{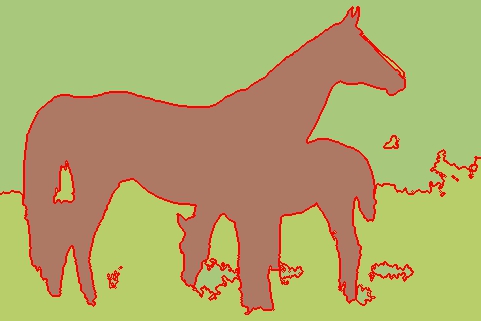}}
\end{minipage}
\begin{minipage}[b]{0.49\linewidth}
  \centerline{\includegraphics[width=\textwidth]{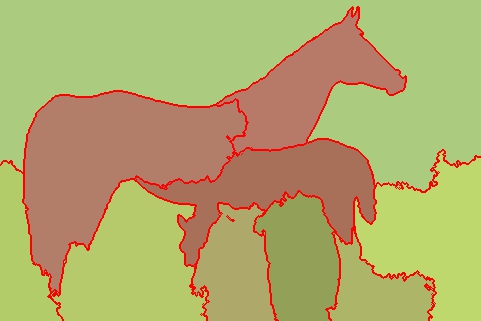}}
\end{minipage}
\begin{minipage}[b]{0.49\linewidth}
  \centerline{(c)}
\end{minipage}
\begin{minipage}[b]{0.49\linewidth}
  \centerline{(d)}
\end{minipage}
\begin{minipage}[b]{0.49\linewidth}
  \centerline{\includegraphics[width=\textwidth]{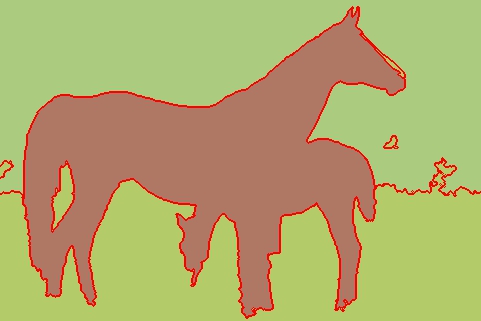}}
\end{minipage}
\begin{minipage}[b]{0.49\linewidth}
  \centerline{\includegraphics[width=\textwidth]{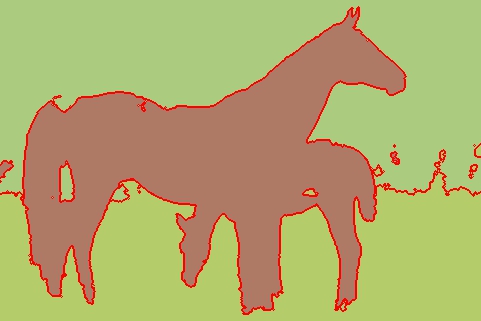}}
\end{minipage}
\begin{minipage}[b]{0.49\linewidth}
  \centerline{(e)}
\end{minipage}
\begin{minipage}[b]{0.49\linewidth}
  \centerline{(f)}
\end{minipage}
\begin{minipage}[b]{0.49\linewidth}
  \centerline{\includegraphics[width=\textwidth]{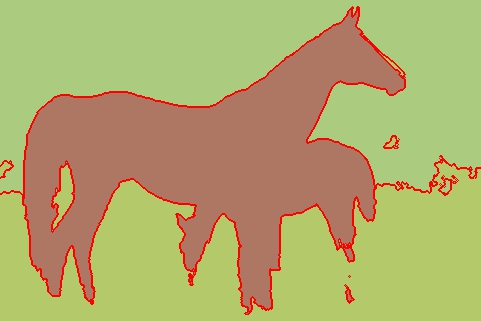}}
\end{minipage}
\begin{minipage}[b]{0.49\linewidth}
  \centerline{\includegraphics[width=\textwidth]{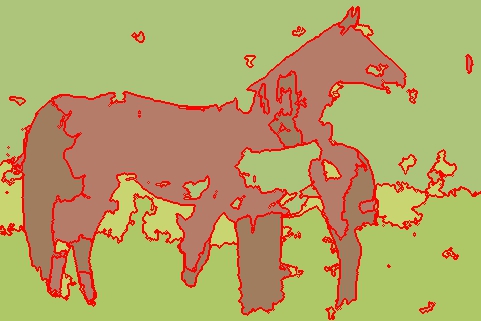}}
\end{minipage}
\begin{minipage}[b]{0.49\linewidth}
  \centerline{(g)}
\end{minipage}
\begin{minipage}[b]{0.49\linewidth}
  \centerline{(h)}
\end{minipage}
\caption{Influence of different covariance descriptors to RBFLE and LRR: (a) original image; (b) ground truth; (c) result of CovI + RBFLE; (d) result of CovI + LRR; (e) result of CovII + RBFLE; (f) result of CovII + LRR; (g) result of CovIII + RBFLE; (h) result of CovIII + LRR.}
\label{fig:visualCompare1}
\end{figure}

\begin{figure}[H]
\begin{minipage}[b]{0.49\linewidth}
  \centerline{\includegraphics[width=\textwidth]{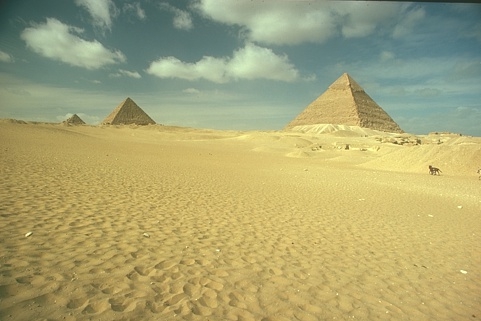}}
\end{minipage}
\begin{minipage}[b]{0.49\linewidth}
  \centerline{\includegraphics[width=\textwidth]{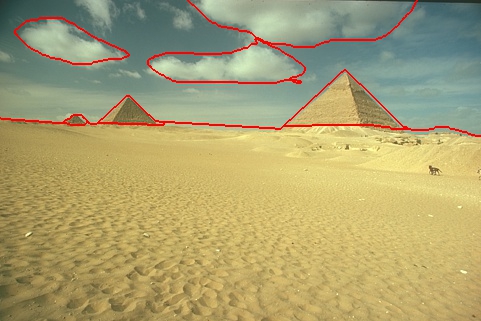}}
\end{minipage}
\begin{minipage}[b]{0.49\linewidth}
  \centerline{(a)}
\end{minipage}
\begin{minipage}[b]{0.49\linewidth}
  \centerline{(b)}
\end{minipage}
\begin{minipage}[b]{0.49\linewidth}
  \centerline{\includegraphics[width=\textwidth]{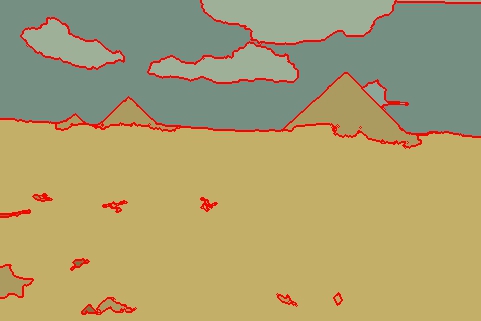}}
\end{minipage}
\begin{minipage}[b]{0.49\linewidth}
  \centerline{\includegraphics[width=\textwidth]{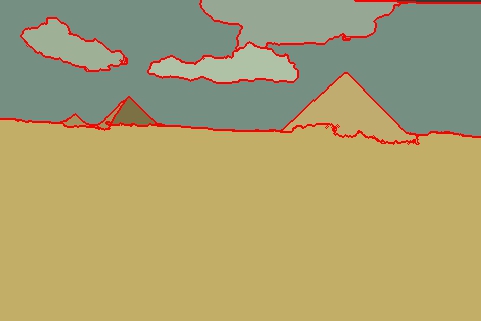}}
\end{minipage}
\begin{minipage}[b]{0.49\linewidth}
  \centerline{(c)}
\end{minipage}
\begin{minipage}[b]{0.49\linewidth}
  \centerline{(d)}
\end{minipage}
\begin{minipage}[b]{0.49\linewidth}
  \centerline{\includegraphics[width=\textwidth]{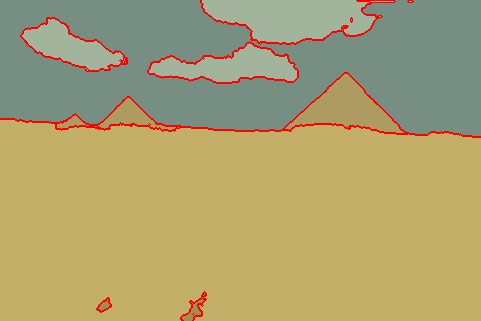}}
\end{minipage}
\begin{minipage}[b]{0.49\linewidth}
  \centerline{\includegraphics[width=\textwidth]{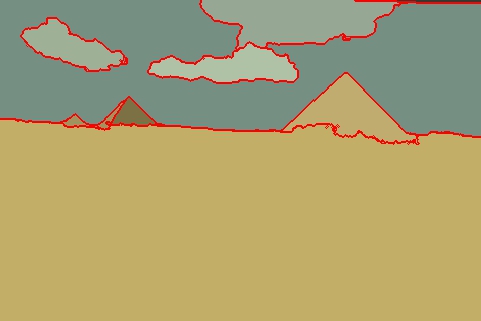}}
\end{minipage}
\begin{minipage}[b]{0.49\linewidth}
  \centerline{(e)}
\end{minipage}
\begin{minipage}[b]{0.49\linewidth}
  \centerline{(f)}
\end{minipage}
\begin{minipage}[b]{0.49\linewidth}
  \centerline{\includegraphics[width=\textwidth]{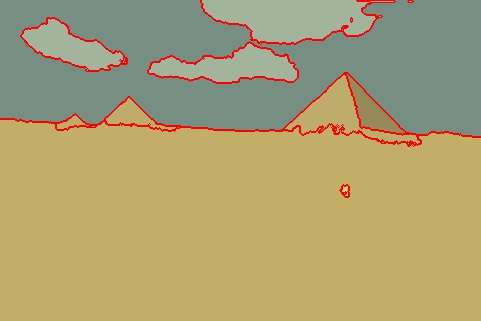}}
\end{minipage}
\begin{minipage}[b]{0.49\linewidth}
  \centerline{\includegraphics[width=\textwidth]{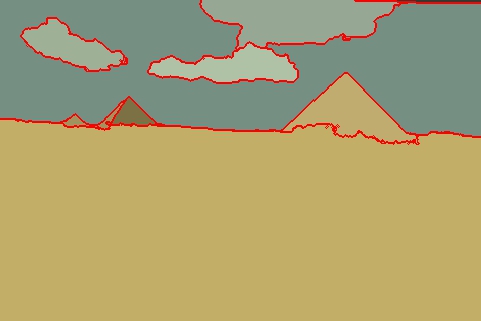}}
\end{minipage}
\begin{minipage}[b]{0.49\linewidth}
  \centerline{(g)}
\end{minipage}
\begin{minipage}[b]{0.49\linewidth}
  \centerline{(h)}
\end{minipage}
\caption{Another example: (a) original image; (b) ground truth; (c) result of CovI + RBFLE; (d) result of CovI + LRR; (e) result of CovII + RBFLE; (f) result of CovII + LRR; (g) result of CovIII + RBFLE; (h) result of CovIII + LRR.}
\label{fig:visualCompare2}
\end{figure}

In addition, Figure.\ref{fig:visualCompare1} and Figure.\ref{fig:visualCompare2} display some results of the algorithms with different covariance descriptors. In Figure.\ref{fig:visualCompare1}, the chaos appears in both algorithms when the covariance descriptor becomes more complicated. While in Figure.\ref{fig:visualCompare2}, the performance of the algorithms benefits from the dimensionality increasing of the covariance descriptors.

\section{Conclusion}
\label{conclusion}
The multi-collinearity usually happens when constructing covariance descriptors. It brings redundancy and noise into the covariance descriptors, which can distort the true data. 

We present two approaches for reducing the effect of multi-collinearity. One is to measure the distance between covariance descriptors by a RBF kernel with a particular geodesic distance; another is to apply low rank representation algorithm on the Riemannian manifold. Our empirical experiments show that LRR method is good for covariance matrix based segmentation since it captures the subspace structure of the data set which is less effected by the noises.

However, the LRR algorithm in this paper is based on an extrinsic metric of the manifold. In the future, we will explore the LRR algorithm with the intrinsic properties of the manifold, i.e. by a geodesic distance.

%
\bibliographystyle{abbrv}
\bibliography{ImageSegmentationWithSuperpixelBasedCovarianceDescriptor}  
%
%

\appendix 
The solution of Eq.(\ref{LRR}) is partly referred to the work of Wang et al.~\cite{wang2015low}, but the distance induced by {\em Frobenius} norm is not geodesic. The problem is rephrased as follows.

Find a matrix $Z$ that satisfied,
\begin{equation}
\begin{array}{ll}
&\min_{E,Z}{\|E\|_F^2 + \lambda \|Z\|_{\ast}},\\
 &s.t. \quad\mathcal{X} = \mathcal{X}_{\times _3}Z+E
\end{array}
\end{equation}
where $\mathcal{X}$ is a 3-order tensor stacking by covariance matrices $(X_i)_{d\times d}$, $i=1,2,...,n$; $\|\cdot\|_F$ is the $Frobenius$ norm; $\|\cdot\|_{\ast}$ is the nuclear norm; $\lambda$ is the balance parameter; $\times_3$ means mode-3 multiplication of a tensor and matrix~\cite{kolda2009tensor}. 

For the error term $E$, we have $\|E\|_F^2 = \|\mathcal{X}-\mathcal{X}_{\times 3}Z\|_{F}^2$, and we can rewrite $\|E\|_F^2$ as,
\begin{equation}
\|E\|_F^2 = \sum_{i}^{N}\|E_i\|_F^2,
\end{equation}  
where $E_i = X_i - \sum_j^Nz_{ij}X_j $, i.e. the $i$-$th$ slice of $E$.

Note that for matrix $A$, it holds $\|A\|_{F}^2 = tr(A^TA)$, and $X_i$ is symmetric, so, the above equation can be expanded as,
\begin{equation}
\begin{array}{ll}
&\|E_i\|_F^2 = tr[(X_i - \sum_j^Nz_{ij}X_j)^T(X_i-\sum_j^Nz_{ij}X_j)]\\
&=tr(X_i^TX_i) - tr(X_i^T\sum_j^Nz_{ij}X_j) -tr(\sum_j^Nz_{ij}X_j^TX_i)\\
&+tr(\sum_{j_1}^Nz_{ij_1}X_{j_1}^T\sum_{j_2}^Nz_{ij_2}X_{j_2})\\
&=tr(X_iX_i) - 2tr(\sum_{j}^Nz_{ij}X_iX_j) + tr(\sum_{j_1,j_2}^Nz_{ij_1}z_{ij_2}X_{j_1}X_{j_2}).
\end{array}
\end{equation} 

Let $\Delta$ be a symmetric matrix of size $N\times N$, whose entries are $\Delta_{ij}=\Delta_{ji}=tr(X_iX_j)$. Because $X_i$ is a symmetric matrix,  $\Delta_{ij}$ can be written as $\Delta_{ij}=vec(X_i)^Tvec(X_j)$, where $vec(\cdot)$ is an operator that vectorized a matrix. As a Gram matrix, $\Delta$ is positive semidefinite. So, we have,
\begin{equation}
\begin{array}{ll}
\|E_i\|_F^2 &= \Delta_{ii} - 2\sum_{j=1}^{N}z_{ij}\Delta_{ij} + \sum_{j_1}^{N}\sum_{j_2}^{N}z_{ij_1}z_{ij_2}\Delta_{j_1j_2}\\
&= \Delta{ii} - 2\sum_{j=1}^{N}z_{ij}\Delta_{ij} + \textbf{z}_i\Delta \textbf{z}_i^T.
\end{array}
\end{equation}
For $\Delta = PP^T$, 
\begin{equation}
\begin{array}{ll}
\|E\|_F^2 &= \sum_{i=1}^{N}\Delta_{ii} - 2tr[Z\Delta] + tr[Z\Delta Z^T]\\
&= C + \|ZP-P\|_{F}^2.
\end{array}
\end{equation}
Then, the optimization is equivalent to:
\begin{equation}
\min_{Z}\|ZP-P\|_F^2 +\lambda\|Z\|_{\ast}.
\end{equation}

Let $\Delta$ be a symmetric matrix, whose entries are $\Delta_{ij}=\Delta_{ji}=tr(X_iX_j)$, and $P = \Delta^{\frac{1}{2}}$. First, we transform the above equation into an equivalent formulation 
\begin{equation}
\begin{array}{ll}
&\min_{Z}\frac{1}{\lambda}\|ZP-P\|_{F}^2 +\|J\|_{\ast},\\
 &s.t. \qquad  J=Z.
\end{array}
\end{equation}
Then by ALM, we have,
\begin{equation}
\min_{Z,J}\frac{1}{\lambda}\|ZP-P\|_{F}^2 +\|J\|_{\ast} + <Y,Z-J>+\frac{\mu}{2}\|Z-J\|_{F}^2,
\end{equation} 
where $Y$ is the Lagrange coefficient, $\lambda$ and $\mu$ are scale parameters. 

The above problem can be solved by the following two subproblems~\cite{lin2010augmented},
\begin{equation}
J_{k+1}=\min_{J}(\|J\|_{\ast}+<Y,Z_k-J> +\frac{\mu}{2}\|Z_k-J\|_F^2)
\end{equation}  
and,
\begin{equation}
Z_{k+1} = \min_{Z}(\frac{1}{\lambda}\|ZP-P\|_F^2+<Y,Z-J_k>+\frac{\mu}{2}\|Z-J\|_F^2).
\end{equation}
Fortunately according to~\cite{cai2010singular}, the solutions for the above subproblems have the following close forms,
\begin{equation}
J = \Theta(Z+\frac{Y}{\mu}), 
\end{equation}
\begin{equation}
Z = (\lambda\mu J -\lambda Y +2\Delta)(2\Delta +\lambda\mu I)^{-1},
\end{equation}
where $\Theta(\cdot)$ is the singular value thresholding operator~\cite{cai2010singular}.

Thus, by iteratively updating $J$ and $Z$ until the converge conditions are satisfied, a solution for Eq.(\ref{LRR}) can be found.

\end{document}